\documentclass[10pt,twocolumn,letterpaper]{article}

\usepackage[pagenumbers]{cvpr} 
\usepackage{tabularx} 

\definecolor{cvprblue}{rgb}{0.21,0.49,0.74}
\usepackage[pagebackref,breaklinks,colorlinks,allcolors=cvprblue]{hyperref}

\def\confName{CVPR}
\def\confYear{2025}

\title{Phys4DGen: Physics-Compliant 4D Generation with Multi-Material Composition Perception}

\author{
Jiajing Lin \quad
Zhenzhong Wang \quad
Dejun Xu \quad
Shu Jiang \quad
Yunpeng Gong \quad
Min Jiang \thanks{Corresponding author: Min Jiang, minjiang@xmu.edu.cn} \\
School of Informatics, Xiamen University
}

\begin{document}

\newcommand{\lin}[1]{\textcolor{blue}{#1}}

\maketitle
\begin{abstract}
4D content generation aims to create dynamically evolving 3D content that responds to specific input objects such as images or 3D representations. Current approaches typically incorporate physical priors to animate 3D representations, but these methods suffer from significant limitations: they not only require users lacking physics expertise to manually specify material properties but also struggle to effectively handle the generation of multi-material composite objects. To address these challenges, we propose Phys4DGen, a novel 4D generation framework that integrates multi-material composition perception with physical simulation. The framework achieves automated, physically plausible 4D generation through three innovative modules: first, the 3D Material Grouping module partitions heterogeneous material regions on 3D representations' surfaces via semantic segmentation; second, the Internal Physical Structure Discovery module constructs the mechanical structure of object interiors; 
finally, we distill physical prior knowledge from multimodal large language models to enable rapid and automatic material properties identification for both objects' surfaces and interiors. 
Experiments on both synthetic and real-world datasets demonstrate that Phys4DGen can generate high-fidelity 4D content with physical realism in open-world scenarios, significantly outperforming state-of-the-art methods.
\end{abstract}

\begin{figure}[htbp]
	\centering
	\includegraphics[width=1\linewidth]{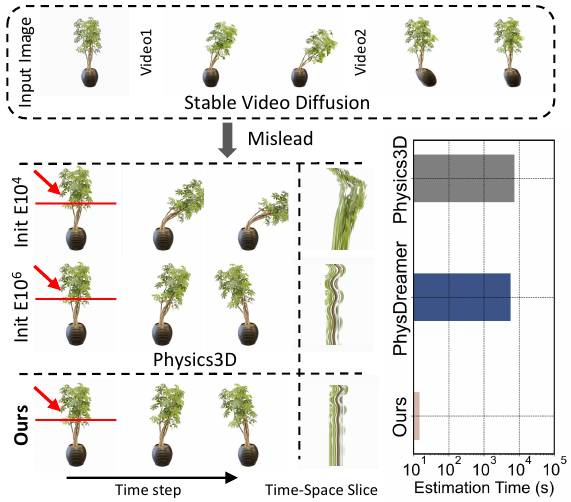}
	\caption{
    The red arrows indicate the direction of external forces. 
    We use the space-time slice (right column), where the vertical axis represents time and the horizontal axis shows a spatial slice of the object (marked by red lines), to reveal motion intensity and frequency.
    As shown in the top, diffusion models embed unrealistic motion priors that may mislead the estimation process—e.g., Physics3D consistently overestimates the softness of the ficus, deviating from physical plausibility. 
    Additionally, the accuracy of such approaches heavily depends on the setting of initial material properties (e.g., Init Young's modulus $10^4$ vs. $10^6$). 
    In contrast, our method achieves more accurate material properties estimation within 14.88 seconds, enabling reliable simulation.}
	\label{fig:intro}
\end{figure}

\section{Introduction}

The creation of 4D content has gained substantial importance across multiple domains, including computer animation, interactive gaming, and immersive virtual reality applications~\cite{Art_Create}. In particular, recent advances in generative models have fundamentally transformed 4D generation due to their powerful visual priors~\cite{DM-image-1, DM-image-2, DM-video-1, DM-video-2, DreamFusion, DreamGaussian}. They leverage dynamic priors from video diffusion models~\cite{Animate124, DreamGaussian4D, 4Diffusion, STAG4D, EG4D}, enabling the automated production of high-quality 4D content. However, these data-driven approaches fundamentally lack physical modeling constraints, often resulting in generated motions that violate physical laws and exhibit noticeable inconsistencies. 

To ensure the generation of physically realistic 4D content, recent works~\cite{Physgaussian, Spring-Mass, PAC-NeRF} have explored the incorporation of physical priors, such as continuum mechanics, to animate 3D representations~\cite{NeRF, 3DGS}.
PhysGaussian~\cite{Physgaussian} pioneered the integration of physical properties into 3D Gaussian Splatting (3DGS)~\cite{3DGS} representations, and introduced the Material Point Method (MPM)~\cite{MPM-1} for dynamic generation. However, it requires manually setting the material type and properties of the simulation object.
Methods such as PAC-NeRF~\cite{PAC-NeRF} and GIC~\cite{GIC} are capable of estimating physical properties under the supervision of multi-view videos. However, they rely heavily on multi-view video data, which is often difficult to obtain, thereby significantly hindering their practical applicability.
PhysDreamer~\cite{PhysDreamer}, DreamPhysics~\cite{DreamPhysics}, and Physics3D~\cite{Physics3D} leverage dynamic priors from pre-trained video diffusion models to guide the estimation of material properties, based on a given 3DGS model rather than multi-view videos. This enables automatic material properties determination without strict input constraints. 

Despite these advances, existing methods still face several critical challenges. 
1) First, these methods typically assume that objects are uniform entities made of a single material, whereas real-world objects often consist of multiple heterogeneous materials.
Failing to distinguish between different materials hinders the accurate simulation of localized deformation behaviors, reducing physical realism.
2) Second, while 3DGS effectively captures the surface geometry of objects, it lacks the capability to model internal structures.
However, object interiors often contain multiple materials, which may even differ from those on the surface.
Simulations based on such structure-agnostic representations are prone to structural collapse 
under large deformations.
3) In addition, as shown in Fig. \ref{fig:intro}, the dynamic priors embedded in video diffusion models, which lack physical constraints, may mislead the estimation of material properties.
Furthermore, due to their iterative optimization process, these methods are computationally expensive.
Both their convergence speed and estimation quality are also highly sensitive to initial material properties settings, which typically require domain-specific physics knowledge. This poses a significant barrier for general users who lack such expertise in 4D content generation.

To overcome the limitations of previous approaches, we introduce {\em Phys4DGen}, a novel physics-driven 4D generation framework that integrates multi-material composition perception into the 4D generation pipeline for the first time, enabling fast, user-friendly, and physically plausible 4D content generation from a single image or 3D representation. Our framework addresses three key challenges: (1) the 3D Material Grouping module, which extends the segmentation semantics of large vision models (e.g., SAM2\cite{SAM2}) from 2D to 3D space for accurate surface material grouping; (2) the Internal Physical Structure Discovery module, which models the mechanical structure of object interiors; and (3) a multimodal physics expert that leverages physical knowledge embedded in large language models to automatically and rapidly identify material properties for both surfaces and internal structures. 
By unifying these components, {\em Phys4DGen} enables more complete and physically realistic 4D generation. Our key contributions include:
\begin{itemize}
\item We propose a 4D generation framework that distills prior knowledge from a foundation model to enable the multi-material composite perception, while incorporating physical simulations to achieve user-friendly and physically realistic 4D generation.
\item 3D Material Grouping is introduced to partition object surfaces into distinct material regions, and Physical Internal Structure Discovery for modeling internal structures, together enabling the handling of multi-material composition.
\item We are the first to leverage physical prior knowledge from a multimodal large language model to enable automatic and efficient material identification.
\item Extensive qualitative and quantitative comparisons on both synthetic and real-world datasets demonstrate that our method can generate physically consistent and high-fidelity 4D content across various materials.
\end{itemize}

\section{Related Work}
\label{sec:related_work}
\subsection{4D Generation} 
4D generation aims to generate dynamic 3D content that aligns with user input conditions such as text, images, and videos. 
Unlike 3D generation~\cite{3DGeneration-image-1,3DGeneration-image-2,3DGeneration-image-3, 3DGeneration-image-4,3DGeneration-image-5,3DGeneration-image-6}, which primarily focuses on producing spatially consistent geometry and appearance, 4D generation must additionally ensure temporal realism and consistency across frames, making the task significantly more challenging. 
Based on the input conditions, 4D generation can be categorized into three types: text-to-4D~\cite{Nvidia, 4D-FY, TC4D, MAV3D, AYG}, video-to-4D~\cite{4DGen, Dreamscene4d, Gaussianflow, Efficient4D, 4Diffusion}, and image-to-4D~\cite{Animate124, Diffusion2, EG4D, SV4D}.  
MAV3D ~\cite{MAV3D} first employs temporal SDS from the text-to-video diffusion model to optimize HexPlane~\cite{Hexplane} representation. 
4D-fy~\cite{4D-FY} introduces hybrid score distillation during training for high-quality text-to-4D generation. 
Instead of text input, 
Consistent4D~\cite{Consistent4D} uses SDS for geometry optimization and interpolation loss for spatiotemporal consistency in 4D generation from monocular video. 
4DGen~\cite{4DGen} supervised by pseudo labels generated from multi-view diffusion model. 
Animate124~\cite{Animate124} pioneered an image-to-4D framework using a coarse-to-fine strategy that combines different diffusion priors~\cite{lu2023tf,lu2024mace}. 
Generating 4D content from an image in DreamGaussian4D ~\cite{DreamGaussian4D} avoids using temporal SDS and instead performs optimization based on reference videos generated by a video diffusion model. 
Our framework can generate physically plausible and temporally coherent 4D content efficiently, without extra optimization steps.

\begin{figure*}[htbp]
	\centering
	\includegraphics[width=1\linewidth]{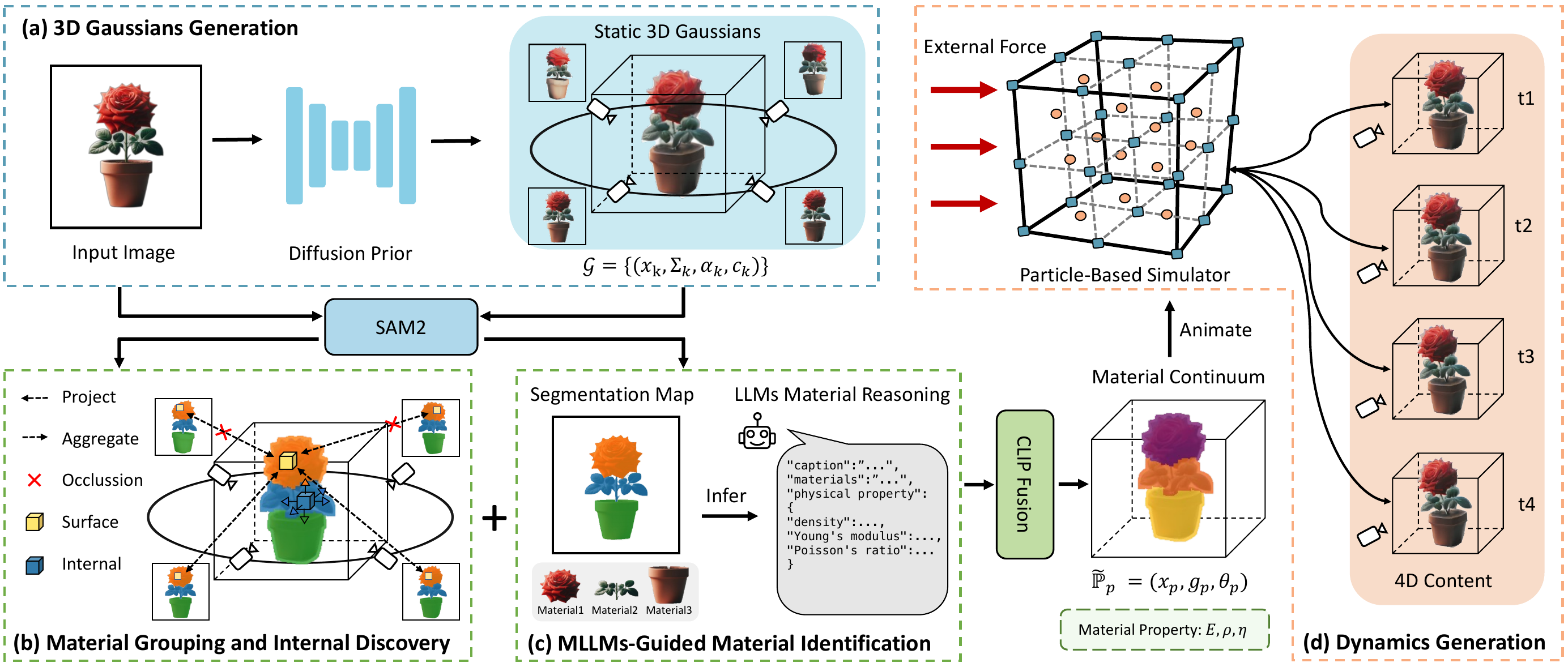}
	\caption{\textbf{Framework of \textit{Phys4DGen}}. 
(a) \textbf{3D Gaussians Generation}: Given an input image, a static 3D Gaussians is generated under the guidance of the diffusion model.
(b) \textbf{Material Grouping and Internal Discover}: 3D Material Grouping is applied to partition the 3D Gaussians into distinct material groups. Concurrently, Internal Physical Structure Discovery is used to fill internal particles and determine their corresponding material groups. 
(c) \textbf{MLLMs-Guided Material Identification}: Surface and internal material properties are visually inferred by MLLMs. These inferred results are then integrated into the 3D representation $\mathbb{P}$ through the CLIP Fusion module, forming a material continuum representation $\tilde{\mathbb{P}}$.
(d) \textbf{4D Dynamics Generation}: Given external forces, MPM simulation is performed to animate the material continuum, thereby generating 4D content.
}
	\label{fig:main_pipe}
\end{figure*}

\subsection{Physics-Based Dynamic Generation}
Some recent methods have attempted to leverage physical simulation to create visual dynamics with physical realism. 
PhysGen~\cite{PhysGen} introduces rigid-body physics simulation to achieve physically grounded video synthesis from a single image.
PAC-NeRF~\cite{PAC-NeRF} recovers object geometry and physical properties from multi-view videos by combining NeRF with differentiable physics, without requiring known shapes. 
However, NeRF’s implicit representation is not ideal for physical simulation.
The explicit representation of 3DGS~\cite{3DGS, wu2025textsplat} through a set of anisotropic Gaussian kernels, which can be interpreted as particles in space, enables the many applications of physical simulations~\cite{Physgaussian, Fluid, Spring-Mass, lin2024phy124}. 
PhysGaussian~\cite{Physgaussian} is the first to apply MPM to 3D Gaussian representations, enabling the simulation of realistic physical dynamics. 
Spring-Mass~\cite{Spring-Mass} introduces a novel integration of a spring-mass system into the 3DGS framework for elastic material simulation.
However, they require manual specification of material types, material properties, and simulation regions.
PhysDreamer~\cite{PhysDreamer} utilizes reference videos from video diffusion models for supervision to estimate material properties. 
Likewise, DreamPhysics and Physics3D~\cite{DreamPhysics, Physics3D} use score distillation sampling from video diffusion models to optimize physical properties. 
The stochastic nature of video diffusion models and their non-physical dynamic priors can distort material property estimation. Our method does not require an optimization process and can efficiently infer material properties. Notably, we uniquely consider the possibility that a single object may comprise multiple materials, including differing internal and external compositions.

\section{Methodology}
The proposed \textit{Phys4DGen} is illustrated in Fig \ref{fig:main_pipe}. 
Given a single input image, a static 3D Gaussians representation is generated under the guidance of a diffusion model (Sec. \ref{sec:3D Generation}).
We then employ \textbf{3D Material Grouping} (Sec. \ref{sec:Material Grouping}) to assign suitable material groups to different regions of the 3D Gaussians.
To further explore internal details, \textbf{Internal Physical Structure Discovery} (Sec. \ref{sec:Internal Discovery}) is employed to model internal geometries and assign the internal material groupings.
Subsequently, \textbf{MLLM-Guided Material Identification} (Sec. \ref{sec:Material Identification}) infers both surface and internal material properties for each material region from the input image. 
These are fused into the 3D representation through CLIP Fusion, producing a material continuum representation.
Finally, given external forces and boundary conditions, the 4D dynamics are simulated using MPM (Sec. \ref{sec:4D Generation}) based on the material continuum.

\subsection{Static 3D Gaussians Generation}
\label{sec:3D Generation}
Given the recent advances in image-to-3D generation~\cite{DreamGaussian, LGM}, which have demonstrated the ability to produce high-quality 3D content, we directly employ these methods to generate static 3D Gaussians.
We choose 3DGS as a representation for its explicit representation nature. 3DGS represents 3D objects using a collection of anisotropic Gaussian kernels~\cite{3DGS}, which can be interpreted as particles in space. 
Thus, 3D Gaussians can be viewed as a discretization of the continuum, which is highly beneficial for integrating particle-based physical simulation algorithms.
In this phase, we obtain a static 3D Gaussians $\boldsymbol{\mathcal{G}}$ for subsequent simulation. Each Gaussian kernel can be represented as $\boldsymbol{\mathcal{G}_k} =(\mathbf{x}_k, \mathbf{\Sigma}_k, \alpha_k, \mathbf{c}_k)$. Here, $\mathbf{x}_k$, $\mathbf{\Sigma}_k$, $\alpha_k$, and $\mathbf{c}_k$ represent the center position, covariance matrix, opacity, and color of the Gaussian kernel $k$, respectively.
Additionally, the quality of the 4D content improves with the quality of the static 3D Gaussians generated from the input image. 

\subsection{3D Material Grouping}
\label{sec:Material Grouping}
In practice, many objects are composites composed of different materials. Prior to material identification, it is essential to group objects into distinct material components.
Thus, we propose 3D Material Grouping, which lifts the segmentation semantics from the vision foundation model from 2D to 3D space, enabling the assignment of a unique material group to each Gaussian kernel.

\subsubsection{Pre-Process.}
Given the static 3D Gaussians generated in the previous phase, we render the scene from multiple views, producing an image sequence and its corresponding depth maps.
SAM2~\cite{SAM, SAM2} is a powerful vision foundation model, supports accurate video segmentation. 
To ensure consistency of 2D mask maps across views—i.e., that the same material region receives the same mask index from different views—we consider the multi-view image sequence as a video input and apply SAM2 for segmentation to generate the associated mask maps $\mathbf{M} = \{\mathbf{M}_o\}_{o=1}^N$.

\subsubsection{Projection and Aggregation.} 
We treat each mask index as a material group label $g$, resulting in a sequence of mask maps with consistent material groupings across views.
For a given Gaussian kernel $\mathcal{G}_k \in \mathcal{G}$ and a mask map $\mathbf{M}_o \in \mathbf{M}$, we use the camera’s intrinsic and extrinsic parameters to project the Gaussian kernel into 2D space, obtaining its 2D coordinates $\mathbf{x}_p^{2d}$ on the corresponding mask map. This process can be expressed as:
\begin{equation}
    \mathbf{x}_k^{2d} = \mathbf{K} [\mathbf{R}_o|\mathbf{T}_o] \mathbf{x}_k,
\end{equation}
where $\mathbf{K}$ and $[\mathbf{R}_o|\mathbf{T}_o]$ represent the camera’s intrinsic and extrinsic parameters, respectively. 
We use the 3DGS-estimated depth to check if the Gaussian kernel is visible in the segmentation map $\mathbf{M}_o$. If it is visible, we include the material group from this view in the voting process.
After processing the segmentation maps for all views, we perform majority voting, assigning the material group $g_k$ that appears most frequently across all views to Gaussian kernel $\boldsymbol{\mathcal{G}_k} = {(\mathbf{x}_k, \mathbf{\Sigma}_k, \alpha_k, \mathbf{c}_k, g_k)}$. 
Repeating these steps allows us to determine the material groups for all Gaussian kernels $\mathcal{G}$.

\subsection{Physical Internal Structure Discovery}
\label{sec:Internal Discovery}
Due to an inherent limitation of the 3DGS representation, a large number of Gaussian kernels are distributed only on the surface, leaving the interior empty. This leads to reduced fidelity in physical simulations, especially under large deformations.
To address this limitation, we propose a strategy termed Physical Internal Structure Discovery, which enables internal material filling and grouping.

Concretely, grid particles are initialized by uniform sampling within the bounding box of the 3D Gaussians $\boldsymbol{\mathcal{G}}$.
These particles are then projected onto multi-view mask images and depth maps. Each particle is filtered by comparing the projected depth with the rendered depth and checking whether it lies within the foreground mask. 
Based on the 3D object represented by the valid grid particles, we further classify them into surface and internal particles.

To distinguish surface particles, we analyze the spatial relationship between each Gaussian kernel and the grid particles using the Mahalanobis distance~\cite{MahalanobisDistance}.
Specifically, for a Gaussian kernel with mean $\mathbf{\mu}$ and covariance $\mathbf{\Sigma}$, the Mahalanobis distance from a particle located at position $\mathbf{x}$ is computed as:
\begin{equation}
    D_M(\mathbf{x}) = \sqrt{(\mathbf{x} - \mathbf{\mu})^T \mathbf{\Sigma}^{-1} (\mathbf{x} - \mathbf{\mu})},
\end{equation}
If this distance is less than a predefined threshold, the grid particle is considered to be covered by the Gaussian kernel. All such particles are labeled as surface particles $\mathcal{P}_{S}$, while those not included by any Gaussian are classified as internal particles $\mathcal{P}_{I}$.

Since the internal structure of an object can only be inferred from surface information available in the input image,
we establish a surface-to-interior correspondence by assigning material groups from surface particles to internal ones. 
To ensure closure of the boundary set, we additionally designate the outermost layer of grid particles as surface particles.
Each surface particle is then assigned to its nearest Gaussian kernel via a nearest-neighbor search and inherits its material group accordingly.
For each internal particle, rays are cast along the six principal axes to collect material groups from the intersected surface particles. The most frequently observed material group $g_i$ is then assigned to the internal particle.
Subsequently, we merge 3D Gaussians $\mathcal{G}$ with the filled internal particles $\mathcal{P}_I$ to obtain a unified continuum particles representation with material groups: $\mathbb{P}=\{\mathcal{G} | \mathcal{P}_I\} = \{  \{ (\mathbf{x}_k, g_k)\} | \{(\mathbf{x}_i, g_i)\}\}$.
Based on this representation, we can easily assign material properties to different material regions of the 3D object.

\subsection{MLLMs-Guided Material Identification}
\label{sec:Material Identification}
\subsubsection{Material Information Reasoning} 
In the real world, objects are typically composed of different materials. In 4D generation, users often lack the necessary physical knowledge to provide reasonable material properties for simulation. This greatly limits the physical realism of the simulation results.
Recently, multimodal large language models have advanced rapidly, exhibiting knowledge far beyond that of humans, including rich physical prior knowledge. Inspired by this, we introduce GPT-4o~\cite{GPT4}, which reasons the material properties (e.g., Density $\rho$, Young's modulus $E$, Poisson's ratio $\nu$) of the internal and external parts of objects through vision.

Specifically, the user-input image is treated as the canonical view $\mathbf{I}_c$ for reasoning with the MLLMs. Sub-images representing different material components are extracted from this view based on the segmentation mask map. Following this, the canonical view and its segmented sub-images are fed into GPT-4o, prompting it to reason the internal and external material types and properties for the object described in each segmented sub-image. Detailed prompts are provided in Appendix~\ref{A.2}.
The output format is as follows: \texttt{\{partial object 1: \{surface: \{material type, material property\}, internal: \{material type, material property\}\}, …\}}.
\subsubsection{CLIP Fusion.} To integrate the inferred surface and internal material properties into the 3D representation. 
It is necessary to align the material groupings of the continuum particles—obtained in Sec.~\ref{sec:Material Grouping} and Sec.~\ref{sec:Internal Discovery}—with those derived from the canonical view.
Specifically, we first extract the CLIP embedding for all segmented sub-images in the canonical view and the rendered image $\mathbf{I}_f$ from the front view, which contains the same object information as the canonical view. This is represented as:
\begin{align}
    \mathbf{L}_c(\cdot) &= \mathbf{V}(\mathbf{I}_c \odot \mathbf{M}_c(\cdot)),
    \\
    \mathbf{L}_f(\cdot) &= \mathbf{V}(\mathbf{I}_f \odot \mathbf{M}_f(\cdot)),
\end{align}
where $\mathbf{V}$ is the CLIP image encoder and $\mathbf{L}$ represents the mask CLIP embedding. $\mathbf{M}(\cdot)$ denotes the sub-region of the mask map labeled by the specified mask index.

Next, we calculate the similarity between the CLIP features of the canonical view and those of the rendered image. By selecting the matching pairs with the highest similarity, we establish a correspondence between the material groupings. Based on this, we can assign the surface and internal material information to the corresponding continuum particles. 
Finally, we construct a complete material continuum representation $\tilde{\mathbb{P}} = {(\mathbf{x}_p, g_p, \theta_p)}$, where $\theta_p$ denotes the material properties, for use in subsequent physical simulation to generate 4D dynamics.

\begin{figure*}[htbp]
	\centering
	\includegraphics[width=1\linewidth]{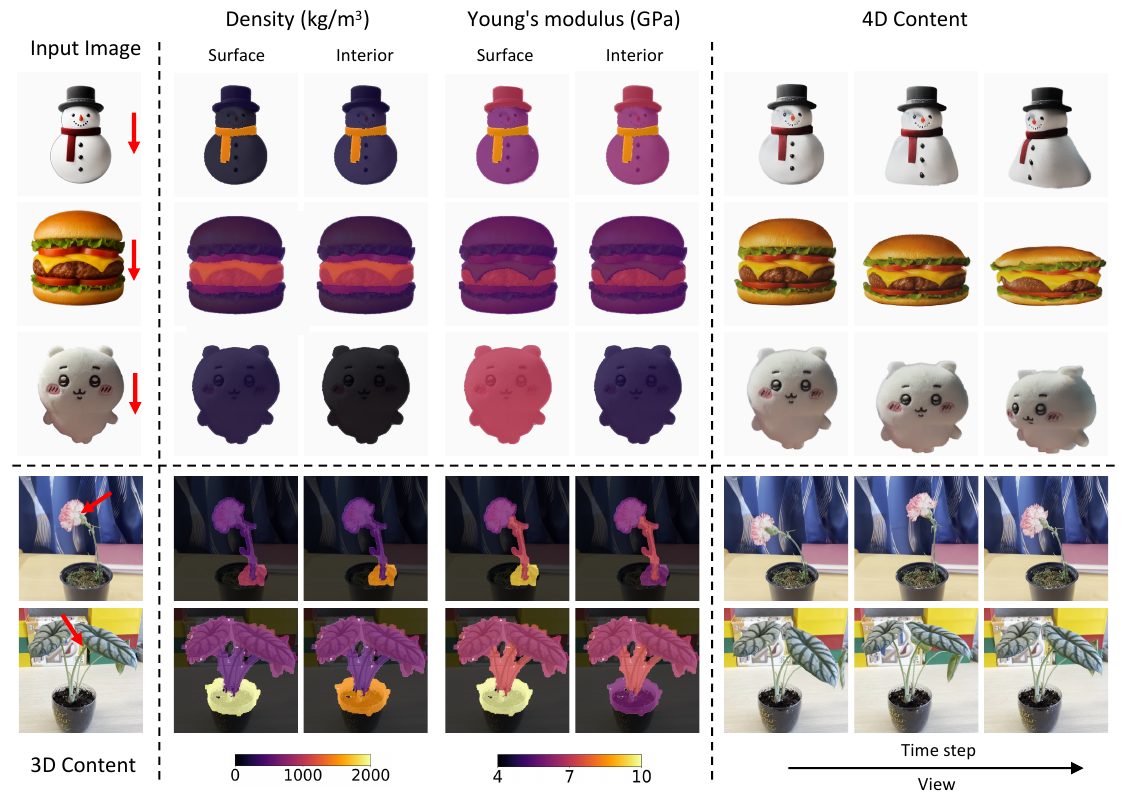}
	\caption{\textbf{Visual results of {\em Phys4DGen}. {\em Phys4DGen} is capable of perceiving the multi-material composition of 3D objects and generating physically realistic 4D content under given external forces (red arrows).}}
	\label{fig:Visulization}
\end{figure*}

\subsection{4D Dynamics Generation}
\label{sec:4D Generation}
\textit{Phys4DGen} can integrate any particle-based physical simulation algorithm. In this paper, we use the Material Point Method (MPM)~\cite{Physgaussian} to simulate the dynamics of 4D content, which enables the modeling of motion and deformation behavior of continuum under external forces. For details on MPM and external forces, please refer to Appendix~\ref{C}.
For physical simulation, we further assign temporal properties $t$ to the material continuum, along with other physical attributes involved in the simulation process, such as mass $m$, deformation gradient $\mathbf{F}$, and velocity $\mathbf{v}$.
Then, we employ MPM to perform physical simulations on the material continuum $\tilde{\mathbb{P}}^t$. This allows us to track the position and local deformation of each particle at every time step:
\begin{equation}
    \mathbf{x}^{t+1}, \mathbf{F}^{t+1}, \mathbf{v}^{t+1} = \text{MPMSimulator}(\tilde{\mathbb{P}}^t),
\end{equation}
where $\mathbf{x}^{t+1} = \{\mathbf{x}_p^{t+1}\}_{p=1}^P$ denotes the positions of all particles at time step $t+1$.
$\mathbf{F}^{t+1} = \{\mathbf{F}_p^{t+1}\}_{p=1}^P$ represents deformation gradients, which describe the local deformation of each particle at time step $t+1$. 
To reconstruct the 3D Gaussian Splatting (3DGS) representation at time step $t$ from the simulation results, we isolate the 3DGS-relevant components from the simulated material continuum. 
To incorporate the local deformation behavior of each GS kernel, we interpret the deformation gradient as a local affine transformation applied to the Gaussian kernel. Consequently, we can derive the covariance matrix of the Gaussian kernel $k$ in step $t+1$:
\begin{equation}
\mathbf{\Sigma}_k^{t+1} = (\mathbf{F}_k^{t+1}) \mathbf{\Sigma}_k^t (\mathbf{F}_k^{t+1})^T.
\end{equation}
At each step of the MPM simulation, we obtain the deformed 3DGS representation. The sequence of 3DGS representations across all time steps collectively forms the 4D content.
This enables the generation of physically plausible 4D dynamics.
Furthermore, since material grouping for 3D objects has already been established in Sec.~\ref{sec:Material Grouping}, external forces can be precisely applied to any specific material region without manual selection, enabling user-friendly interaction and fine-grained control.



\begin{figure*}[htbp]
	\centering
	\includegraphics[width=1\linewidth]{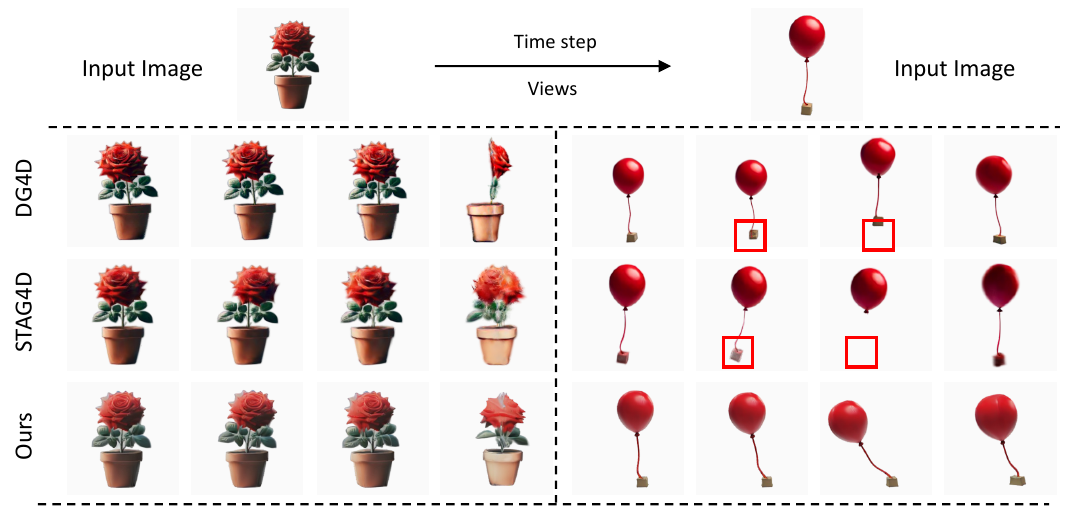}
	\caption{Qualitative comparison in image-to-4D generation. To compare the spatiotemporal consistency, the rendering view changes with each time step. The red box highlights regions exhibiting physically implausible behavior for further observation. The dynamics generated by our method are more consistent with physical laws compared to the baseline method.
    }
	\label{fig:image-to-4D}
\end{figure*}

\section{Experiments}
\subsection{Experimental Setup}
\subsubsection{Implementation Details}
Phys4DGen supports both a single image and a 3D model as input. Given a single image, we use LGM~\cite{LGM} to generate static 3D Gaussians. For material grouping, we render multiview images from the generated 3D Gaussians and apply SAM2~\cite{SAM2} to obtain cross-view consistent material mask maps. We used GPT-4o to identify the material for each material region in the image.
For 4D dynamics generation, we perform physical simulation using MPM~\cite{MPM-1}.
For each example, the simulation environment is configured based on the material information inferred by GPT-4o, and different external forces are applied according to the specific case, allowing the generation of physically plausible dynamic 4D sequences.
All experiments were conducted on NVIDIA A40(48GB) GPU. 
For more detailed information on the experimental settings, please refer to Appendix~\ref{A.1}.

\subsubsection{Datasets}
To thoroughly assess the effectiveness of our approach across varying input types, we establish separate datasets for the Image-to-4D and 3D-to-4D tasks. For the image-to-4D task, we use a total of 11 samples, including 8 synthetic image samples (4 sourced from Zero123\cite{Zero123} and Animate124\cite{Animate124}, and 4 created by ourselves), as well as 3 image samples collected from the real world. 
Following previous work, we employ U2-Net~\cite{U2-Net} to extract the object foreground. 
The datasets feature a range of materials, including elastic, elastoplastic, granular media, and snow.
For the 3D-to-4D task, we choose four real-world static scenes from PhysDreamer~\cite{PhysDreamer} (alocasia, carnations, telephone and hat), along with additional scenes: ficus from PhysGaussain\cite{Physgaussian} and basketball from Physics3D\cite{Physics3D}.

\subsubsection{Baselines}
We compare our method both qualitatively and quantitatively with existing SOTA 4D generation methods. 
For the image-to-4D task, we compare our approach with several image-to-4D generation methods, including DG4D~\cite{DreamGaussian4D}, STAG4D~\cite{STAG4D}, and L4GM~\cite{L4GM}, with a primary focus on assessing their performance in terms of spatiotemporal consistency and physical realism.
For the 3D-to-4D task, we compare our method with PhysGaussian~\cite{Physgaussian}, PhysDreamer~\cite{PhysDreamer}, and Physics3D~\cite{Physics3D}—approaches that incorporate physical priors—using the PhysDreamer datasets. 
The evaluation focuses on two aspects: the physical realism and the efficiency in estimating material properties.

\subsubsection{Metrics}
Following previous works~\cite{Animate124, 4DGen}, we use CLIP-T score which calculates the average cosine similarity between the CLIP embeddings of every two adjacent frames in rendered video from a given view. To further assess the spatiotemporal consistency, we render videos from the right, back, and left views to calculate the CLIP-T-other score. We also conduct a user preference study to evaluate the physical realism (PR) and overall quality (OQ) of the generated 4D content, with both metrics rated on a 10-point scale.
More details on the user study can be found in Appendix~\ref{A.3}

\begin{figure*}[htbp]
	\centering
	\includegraphics[width=1\linewidth]{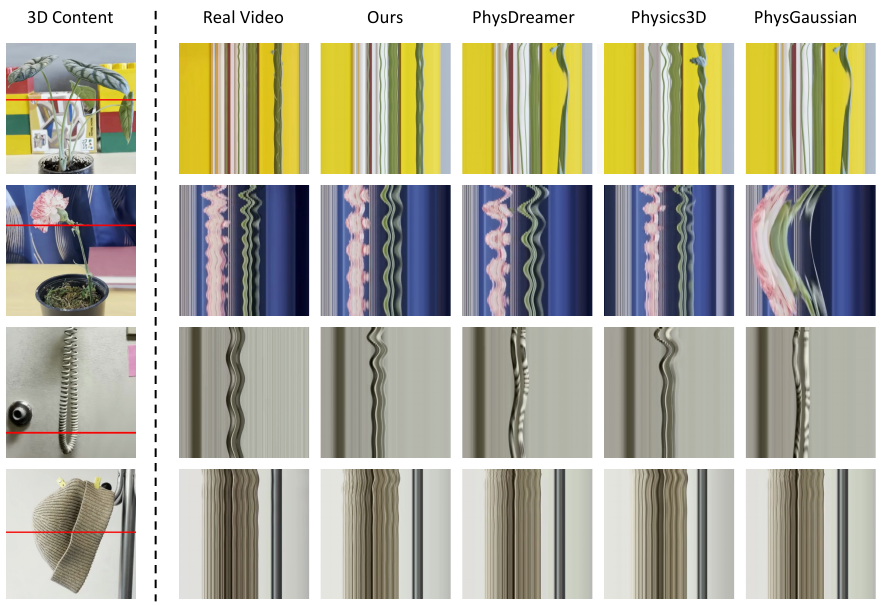}
	\caption{Qualitative comparison for 3D-to-4D generation. We compare our results with real videos and baselines using space-time slices, 
    These slices reveal the motion's intensity and frequency. Our results more closely match the ground truth.}
	\label{fig:3D-to-4D}
\end{figure*}

\subsection{Showcase of 4D Generation Results}
Fig. \ref{fig:Visulization} visualizes the 4D content generated by Phys4DGen. For each example, we present the perceived material properties, including the density and Young's modulus of both the object's surface and internal regions. As shown, Phys4DGen effectively discerns the material compositions of 3D objects, such as differentiating the distinct physical attributes of carnations' petals and stems, and recognizing the material differences in a doll's fabric surface and polyester fiberfill interior. Furthermore, we render the corresponding 4D content from dynamically changing viewpoints. The results demonstrate Phys4DGen's capability to generate physically realistic 4D content from a single image or 3D content.
Please refer to Appendix~\ref{B.1} for additional visual results.

\begin{table}[ht]
\centering
\scalebox{0.85}{
\begin{tabular}{ccccc}
\hline
\textbf{Method} & \textbf{PR} $\uparrow$ & \textbf{OQ} $\uparrow$ &\textbf{CLIP-T} $\uparrow$ & \textbf{CLIP-T other} $\uparrow$ \\ \hline
DG4D~\cite{DreamGaussian4D} & 5.90 & 6.25 & 0.98983 & 0.98536 \\ 
STAG4D~\cite{STAG4D} & 5.55 & 5.97 & 0.98813 & 0.98492 \\
L4GM~\cite{L4GM} & 6.30 & 6.70 & 0.99242 & 0.99275\\ 
Ours & \textbf{7.50} & \textbf{7.72} & \textbf{0.99459} & \textbf{0.99409} \\ \hline
\end{tabular}
}
\caption{Quantitative comparison in image-to-4D generation. PR evaluates the physical realism of the 4D content, OQ measures its overall quality, and CLIP and CLIP-T-other assess its spatiotemporal consistency.
}
\label{tab:image-to-4D}
\end{table}

\subsection{Comparison in Image-to-4D Generation}
Fig. \ref{fig:image-to-4D} presents a qualitative comparison of our Phys4DGen method with state-of-the-art (SOTA) image-to-4D generation methods.
Generating 4D content using STAG4D and DG4D heavily depends on the quality of reference videos produced by video diffusion models. However, due to the inherently stochastic and data-driven nature of these models, it remains challenging to obtain reference videos that simultaneously adhere to physical laws and faithfully align with the user's intent.
This limitation is evident in Fig. \ref{fig:image-to-4D}, where the 4D content generated by STAG4D and DG4D struggles with controllability and often violates fundamental physical principles.
For example, in Fig. \ref{fig:image-to-4D}, the 4D content generated by DG4D~\cite{DreamGaussian4D} for a balloon tied to a wooden block moves upwards, violating gravity.
In stark contrast, Fig. \ref{fig:image-to-4D} clearly demonstrates that our generated 4D content achieves high fidelity and adheres physical laws, significantly outperforming baseline methods. 
This qualitative observation is further supported by the quantitative results presented in Tab.~\ref{tab:image-to-4D}. As shown in the table, our method achieves the highest CLIP-T and CLIP-T-other scores, indicating that it is capable of generating spatiotemporally consistent 4D content.
The user study results on physical realism (PR) and overall quality (OQ) indicate that participants found our generated 4D content more physically plausible and expressed a stronger preference for our results over the baselines.
Additionally, our method offers superior controllability, enabling users to adjust external forces according to their specific needs—an aspect where baseline approaches fall short.

\subsection{Comparison in 3D-to-4D Generation}
Following PhysDreamer\cite{PhysDreamer}, we compare our results with real captured videos and simulations from other methods using space-time slices, where the vertical axis represents time and the horizontal axis shows a spatial slice of the object, as indicated by the red lines in the "object" column. We use 90-frame video sequences to generate space-time slices. This representation enables an effective comparison of the magnitude and frequencies of the oscillatory motions generated by different methods.
Fig. \ref{fig:3D-to-4D} demonstrates that the dynamics generated by our method more closely resemble those of real-world videos. 

\begin{table}[t]
\centering
\scalebox{0.85}{
\begin{tabular}{ccccc}
\hline
\textbf{Method} & \textbf{PR} $\uparrow$ & \textbf{OQ} $\uparrow$ &\textbf{CLIP-T} $\uparrow$ & \textbf{Time} $\downarrow$ \\ \hline
PhysGaussian~\cite{Physgaussian} & 5.67 & 6.30 & 0.99855 & - \\ 
PhysDreamer~\cite{PhysDreamer} & 6.43 & 6.90 & 0.99862 & 5688.79s \\
Physics3D~\cite{Physics3D} & 7.17 & 7.53 & 0.99852 & 7273.32s \\ 
Ours & \textbf{7.87} & \textbf{7.97} & \textbf{0.99925} & \textbf{723.14s+14.88s} \\ \hline
\end{tabular}
}
\caption{Quantitative comparison in 3D-to-4D generation.
Since PhysGaussians lacks a material property estimation stage, we exclude its runtime from our evaluation. Our method requires only 14.88s for property estimation.}
\label{tab:3D-to-4D}
\end{table}

As shown in Tab. \ref{tab:3D-to-4D}, our method achieves the highest average scores in physical realism (PR), overall quality (OQ), and CLIP-T, significantly outperforming baseline methods.
These results validate the effectiveness of our proposed multi-material composition perception capability, which enables relatively accurate material property estimation within only 14.88 seconds, in contrast to the hour-level computation time required by baseline methods. This efficient perception contributes to the generation of 4D content that is not only physically plausible but also more preferred by human observers.
More comparative results are provided in Appendix~\ref{B.2}.

\begin{figure}[t]
	\centering
	\includegraphics[width=1\linewidth]{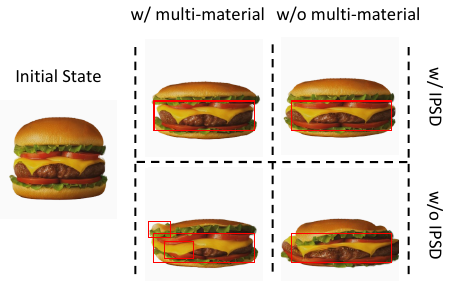}
	\caption{Ablation study on Internal Physical Structure Discover and multi-material partitions through Material Grouping. All examples are tested under the same external forces. The red box is used to assist in observing the deformation.}
	\label{fig:multi_phyfiiling_ablation}
\end{figure}

\subsection{Ablation Analysis}
To validate the effectiveness of Internal Physical Structure Discovery (IPSD) and multi-material partitioning through Material Grouping, we conduct the ablation study illustrated in Fig. \ref{fig:multi_phyfiiling_ablation}.
As shown in the figure, the complete model (top-left) demonstrates ideal simulation performance. With both Material Grouping and IPSD enabled, the bun and the beef patty exhibit different deformation behaviors under the same external force: the bun, being softer, undergoes larger deformation, while the beef, being relatively stiffer, deforms less. Meanwhile, the overall structure remains stable, reflecting strong physical plausibility and internal consistency.
When the Material Grouping module is removed (top-right), internal structural cues are still present, but the model fails to distinguish between materials effectively. As a result, the bun and beef respond with similar levels of deformation under force, which clearly contradicts physical reality.
In contrast, when the IPSD module is removed (bottom-left), the model still captures material differences in deformation response. However, the lack of internal structural support leads to a collapse of the overall geometry under larger external forces, revealing significant issues in maintaining structural stability.
The worst performance occurs when both modules are disabled (bottom-right). The simulation results in chaotic material interactions and complete structural failure.
In summary, both Material Grouping and IPSD are essential for enhancing the physical realism of 4D content generation. They complement each other—Material Grouping ensures that the behavioral differences between materials are properly captured, while IPSD ensures structural integrity—making them indispensable components for achieving faithful and stable physical effects.

\begin{figure}[t]
	\centering
	\includegraphics[width=1\linewidth]{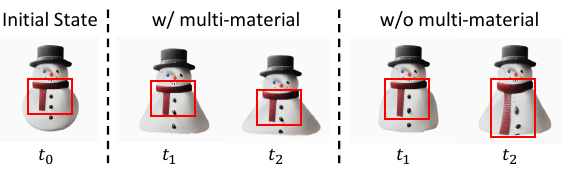}
	\caption{Analysis of material editability enabled by Material Grouping. We are able to simulate objects as a composition of independently controllable sub-material regions.}
	\label{fig:Editable}
\end{figure}

Material Grouping enables material editing by allowing different material parts to be independently controlled during simulation. As shown in Figure \ref{fig:Editable}, we are able to simulate the melting effect specifically on the snowman's body without affecting the rest of the object, such as the scarf. In contrast, without material grouping, the snowman is treated as a uniform material entity. As a result, the melting simulation affects the entire structure, including the scarf, which clearly violates physical plausibility. This demonstrates that multi-material perception is essential for fine-grained control and physically meaningful manipulation.
Additional experimental analyzes can be found in the Appendix~\ref{B.4}.

\section{Conclusion}
In this paper, we propose Phys4DGen, a physics-compliant 4D generation framework that effectively perceives complex multi-material compositions. By seamlessly integrating these perceptual capabilities with physical simulation, our approach enables intuitive and physically plausible 4D generation from a single image or a 3D input.
To handle multi-material compositions, we propose the 3D Material Grouping module, which segments an object surface, represented by 3D Gaussians, into distinct material regions. 
Furthermore, the internal structure of the object is modeled through Physical Internal Structure Discovery. 
We distill extensive physical priors from GPT-4o to identify surface and internal material properties, which are then assigned to the 3D representation to construct a complete simulation object.
Extensive experiments on synthetic and real-world datasets show that our approach generates physically realistic 4D content.
Currently, our method is designed for single-object 4D generation. Extending it to handle multi-object scenarios is an exciting and challenging avenue for future work.

\section{Acknowledgments}
This work was supported by the National Natural Science Foundation of China under Grant No. 62276222.

\clearpage

{
    \small
    \bibliographystyle{ieeenat_fullname}
    \bibliography{main}
}

\clearpage

\appendix
\section*{Overview}
In this appendix, we will provide more implementation details, the design of prompts, and user study details in Sec. \ref{A}. 
Furthermore, we will showcase additional experimental results in Sec. \ref{B}.
In Sec. \ref{C}, we provide preliminaries with 3DGS and a detailed explanation of the derivation process of the MPM algorithm. 
Meanwhile, Our \textbf{source code} and \textbf{rendering video results} are publicly available at \url{https://jiajinglin.github.io/Phys4DGen/}

\section{More Experimental Settings}
\label{A}
\subsection{More Implementation Details}
\label{A.1}
Our method supports 4D generation from either a single image or an given 3D content. For single-image input, we first adopt LGM~\cite{LGM} to generate a static 3DGS representation from the image. For 3D content input, we directly utilize the provided model as the starting point.
For the image-to-4D task, we render 29 images from different viewpoints around the generated 3D representation. 
For the 3D-to-4D task, we render 120 images from randomly sampled viewpoints based on the given 3D Gaussian representation. 
Based on the resulting image sequences, we use SAM2~\cite{SAM2} to extract material grouping masks with cross-view consistency.
We set the occlusion threshold to 0.1. If the difference between the actual and estimated depth is less than this threshold, the Gaussian kernel is considered visible.
For Gaussian kernels occluded from any viewpoint, we assign their material group to that of their nearest neighbor. 
For Gaussians that are occluded from all rendering viewpoints, we assign their material group based on the nearest visible Gaussian in spatial proximity.
For internal physical structure discovery, we voxelize the object's bounding boxes into a 32×32×32 grid. 
We use the GPT-4o to infer material properties and set the inference temperature to 0.8.
After that, we extract the CLIP features of the segmented sub-images using CLIP ViT-B-16~\cite{CLIP} to perform CLIP Fusion.
For all examples in the Physically-Compliant 4D Dynamics 
Generation, the material properties are initialized according to the material properties predicted by the previously described GPT-4o model.
For the image-to-4D task, the foreground region is discretized into a $50^3$ Eulerian grid. We set 714 sub-steps between successive frames, corresponding to a duration of $5 \times 10^{-5}$ seconds for each sub-step, allowing for the generation of 14 frames per second.
For the 3D-to-4D task, the foreground region is discretized into a $64^3$ Eulerian grid. Between successive frames, 400 simulation sub-steps are performed, with the time step set consistent with the image-to-4D setting.

\subsection{Prompting Details}
\label{A.2}
\begin{figure}[htbp]
	\centering
	\includegraphics[width=1\linewidth]{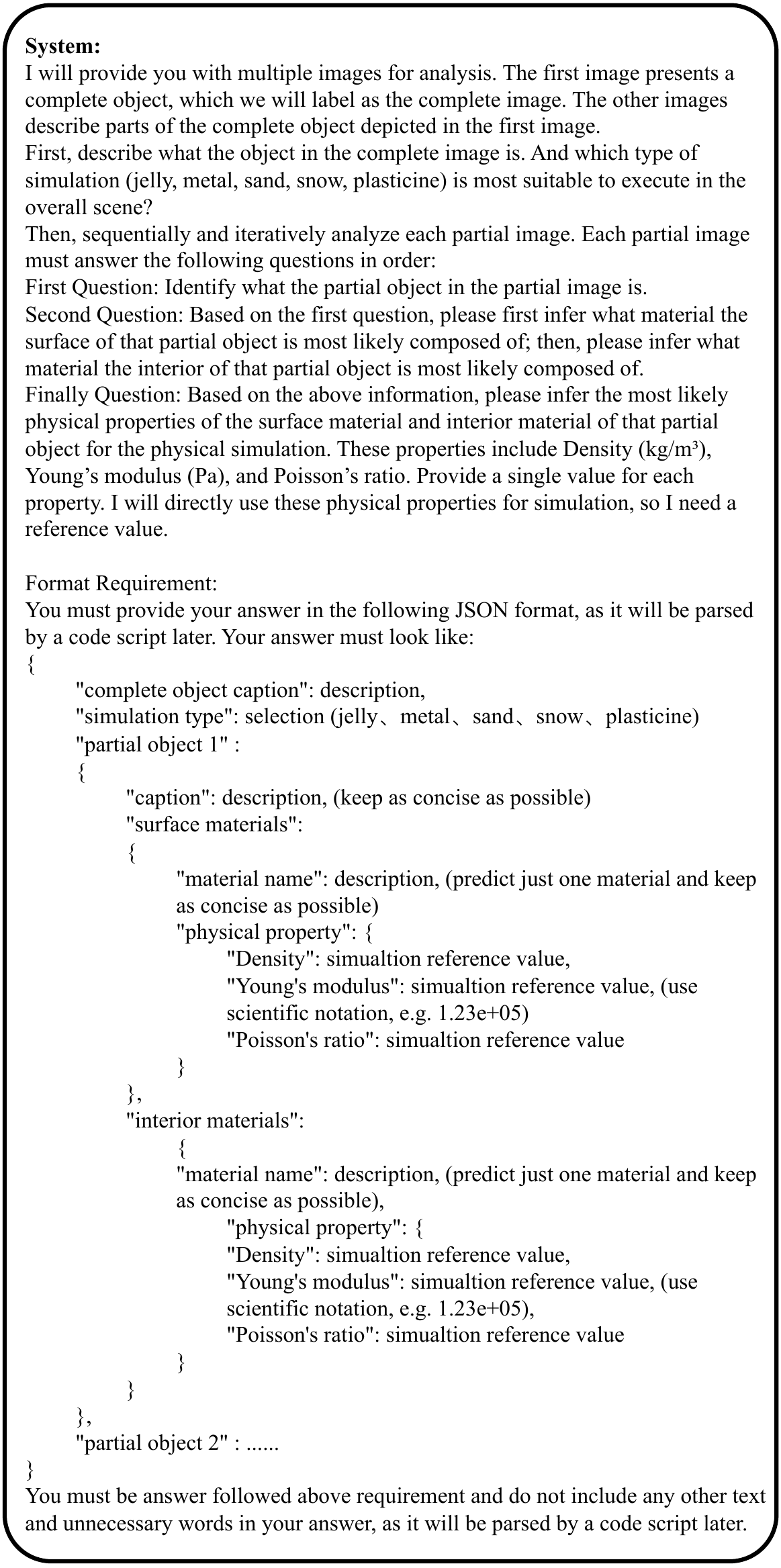}
	\caption{\textbf{Prompt used for material identification}}
	\label{fig:Prompt}
\end{figure}

We provide prompts for material reasoning, as shown in Fig. \ref{fig:Prompt}. By inputting these prompts along with the input image and its segmented sub-images into GPT-4o, we can infer the material type, density, Young's modulus, Poisson's ratio, and other physical properties.

\begin{figure*}[htbp]
	\centering
	\includegraphics[width=1\linewidth]{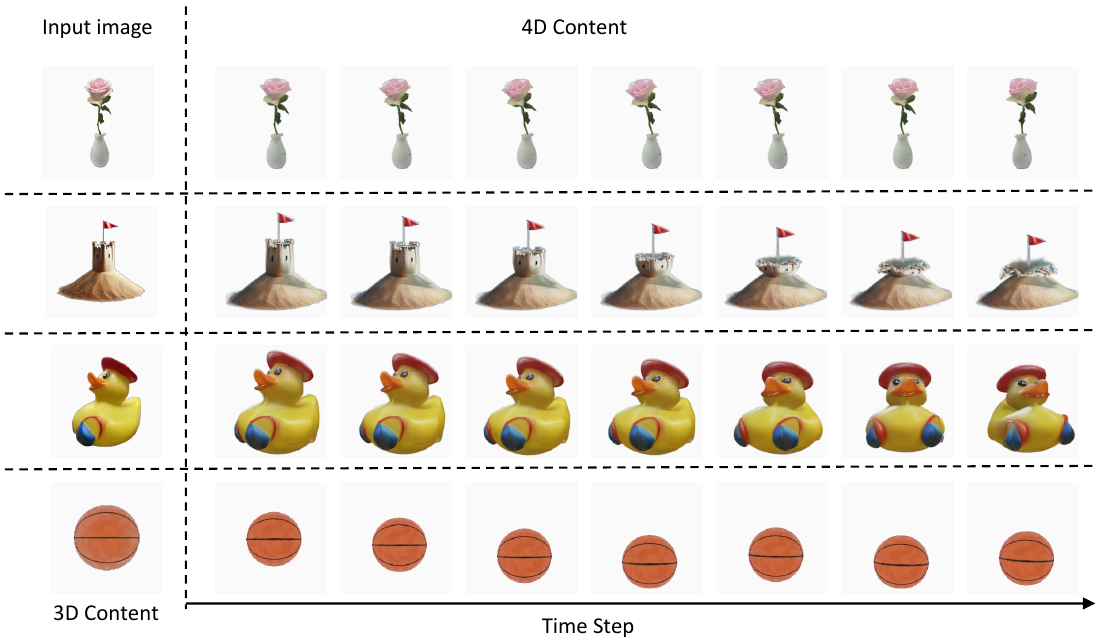}
	\caption{\textbf{Visualization results of more examples.} From the input images, we generated the 4D content, including orange roses swaying in the wind, sandcastle collapsing, and a rubber duck being pressed.}
	\label{fig:more_result}
\end{figure*}

\subsection{User Study Details}
\label{A.3}
To further evaluate the quality of the 4D content generated by our method, we conducted a user study involving 25 volunteer participants.
During the experiment, each participant was asked to view a series of test examples, each generated by multiple existing methods including ours. To eliminate potential ordering bias, the 4D results from different methods were presented in a \textit{randomized sequence} for each example.
Participants independently rated each method on two criteria:
\begin{itemize}
    \item \textbf{Physical Realism (PR)}: Whether the generated motion complies with real-world physical laws, such as whether the deformation aligns with material properties, whether the motion is stable, and whether any physically implausible behaviors occur.

    \item \textbf{Overall Quality (OQ)}: The overall perceptual quality of the animation, including visual coherence, structural consistency, and naturalness of material appearance.
\end{itemize}
Both metrics were rated on a 10-point scale (1 = lowest, 10 = highest). The evaluation was self-paced with no time constraints, and all participants completed the ratings independently to ensure unbiased subjective judgment.

\section{More Experimental Results}
\label{B}
\subsection{Visual Results}
\label{B.1}
We provide visualization results for more examples, as shown in Fig. \ref{fig:more_result}, including pink roses swaying in the wind, a sandcastle collapsing, a rubber duck being pressed and a basketball falling down under gravity.
These results further highlight the generalization ability of our method, showcasing its capability to generate 4D content in diverse real-world scenarios.

\begin{figure*}[htbp]
	\centering
	\includegraphics[width=1\linewidth]{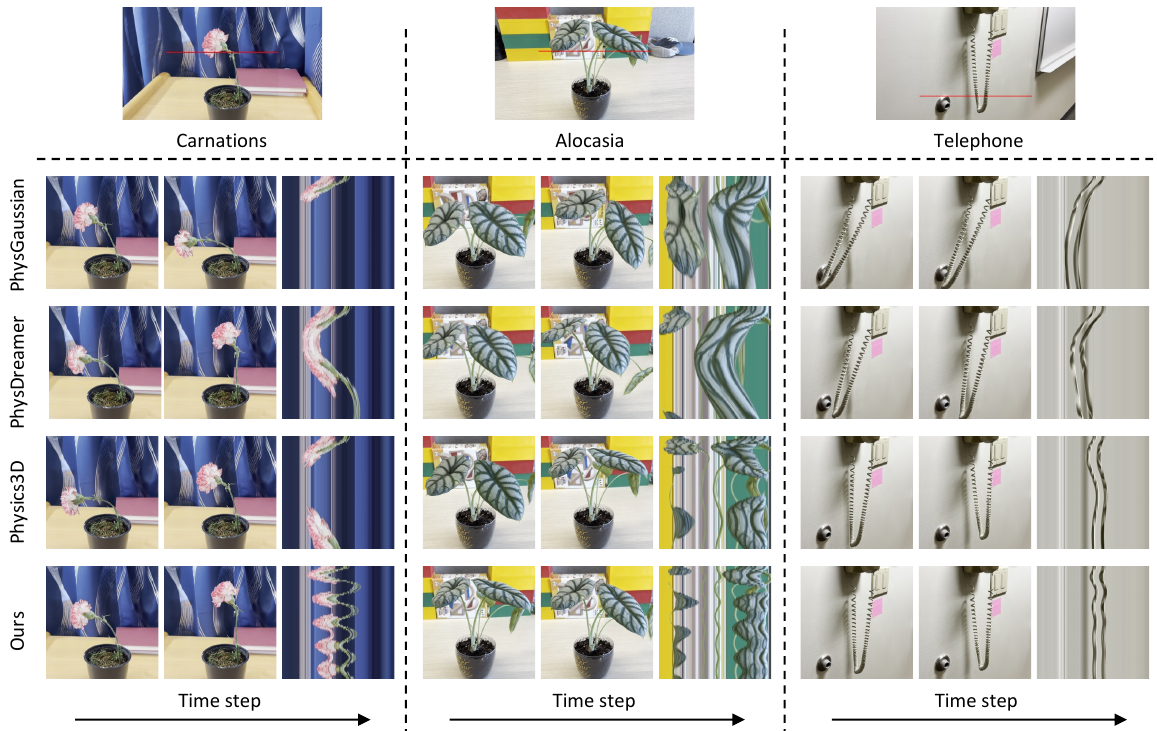}
	\caption{\textbf{More Comparison in 3D-to-4D Generation.} We further tested various methods on the PhysDreamer dataset under large external forces, where our method exhibited dynamics most consistent with physical laws.}
	\label{fig:Appendix_3D-to-4D}
\end{figure*}

\begin{figure*}[htbp]
	\centering
	\includegraphics[width=1\linewidth]{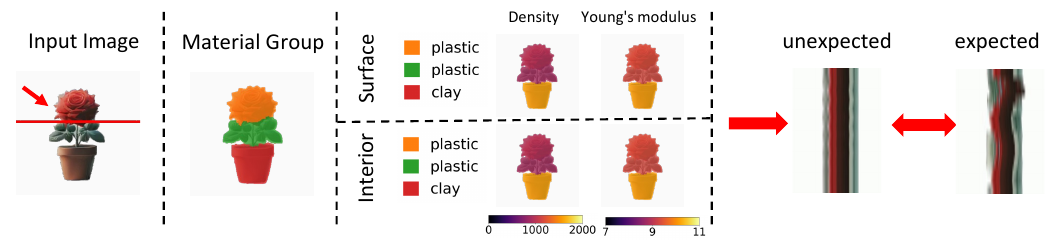}
	\caption{\textbf{Failure Case.} MLLMs misclassified both petals and leaves as plastic and set Young’s modulus to 1 GPa. This led to stiff 4D dynamics that do not match how real flowers bend under force.}
	\label{fig:Appendix_failure_case}
\end{figure*}

\subsection{More Comparisons in 3D-to-4D Generation}
\label{B.2}
To further evaluate the physical realism of different physics-based methods under large external forces, we design three representative experimental scenarios using real-world objects: Carnations, Alocasia, and Telephone. We compare the performance of PhysGaussian, PhysDreamer, Physics3D, and our proposed method. As shown in Fig. \ref{fig:Appendix_3D-to-4D}, we present the corresponding temporal frame sequences along with space-time slices to analyze the oscillation frequency and dynamic response characteristics produced by each method.
Experimental results show that PhysGaussian and PhysDreamer often generate overly distorted and unstable dynamics, lacking physical plausibility. Although Physics3D exhibits a certain degree of elastic recovery, its overall responsiveness remains limited. In contrast, our method demonstrates stronger structural preservation, more realistic material responses, and smoother temporal continuity across diverse scenarios. It is capable of producing physically consistent and stable 4D dynamics for objects with various shapes and material properties.
These results further verify the generalization ability and robustness of our approach under complex external forces, offering a reliable solution for high-fidelity 4D content generation.


\begin{figure}[htbp]
	\centering
	\includegraphics[scale=0.88]{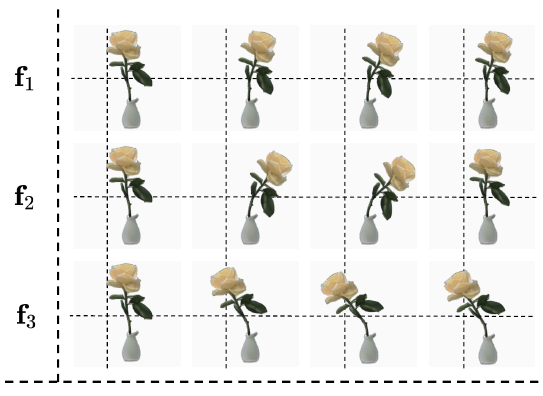}
	\caption{\textbf{External Forces Analysis.} 4D content is generated by applying different external forces $\mathbf{f}$.
    }
	\label{fig:different_force_exp}
\end{figure}

\subsection{Failure Case Analysis}
\label{B.3}
Due to the inherent stochasticity of MLLMs, material properties estimation may occasionally be erroneous. As a failure case, we reference the \textit{Red Flower} example in Fig.~\ref{fig:image-to-4D} of the paper. Due to current review constraints, this visualization cannot yet be shown, but it will be included in the Appendix.
As shown in Fig.~\ref{fig:Appendix_failure_case}, the MLLM incorrectly classified both the petals and leaves as plastic, assigning them a Young’s modulus of 1 GPa. This resulted in a generated 4D dynamics with a rigid appearance, which clearly contradicts the expected soft and flexible behavior of real flowers under force. In contrast, the result shown in Fig.~\ref{fig:image-to-4D} demonstrates a physically consistent 4D dynamics generated based on a correct material prediction. We hypothesize this error was caused by \textbf{visual ambiguity} in the input image: the \textit{Red Flower} appears artificial, leading the MLLM to misinterpret it as a plastic object and overestimate its stiffness.

\subsection{Ablation Studies}
\label{B.4}
\noindent\textbf{The Ability for Fine-grained Controlling 4D Dynamics.}
Fig. \ref{fig:different_force_exp} illustrates the 4D content generated under various external forces, showcasing the fine-grained controllability of \textit{Phys4DGen}. In the first row, external force $\mathbf{f}_1=(1.0,0.0,0.0)$, directed to the right along the x-axis is applied, causing the orange rose to move to the right. In the second row, a larger force $\mathbf{f}_2=(2.0,0.0,0.0)$ is applied in the same direction, resulting in more intense motion.
This demonstrates that \textit{Phys4DGen} can control the strength of motion by adjusting external force. In the third row, a leftward force $\mathbf{f}_3 = (-1.0, 0.0, 0.0)$, with the same magnitude as in the first row, is applied along the x-axis.
As a result, the orange rose moves left under this external force, demonstrating that \textit{Phys4DGen} can control the direction of motion 
To summarize, \textit{Phys4DGen} allows fine-grained control of the dynamics in 4D content by adjusting external forces.

\noindent\textbf{Hyperparameter Analysis for Internal Structure Discovery.} 
During the internal structure discovery stage, we adopted the same set of heuristic-driven parameters across all scenarios, including: grid size = 32 (which controls particle filling density), confidence = 0.95 (used in the chi-squared test function to determine the distance threshold), and max particle number = 10 (the maximum number of particles allowed per Gaussian kernel). 

\begin{figure}[t]
	\centering
	\includegraphics[scale=0.88]{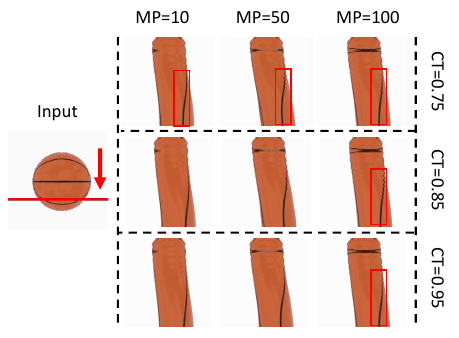}
	\caption{\textbf{Different Settings of Confidence and Max Particles Parameters.} Higher parameters cause more bounces after landing, implying greater elasticity and stiffness.
    }
	\label{fig:Appndix_confidence_and_max_particles}
\end{figure}

\begin{figure}[t]
	\centering
	\includegraphics[scale=0.88]{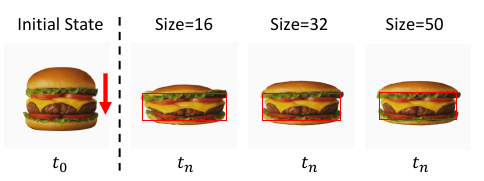}
	\caption{\textbf{Different Settings of Grid Size Parameters.} Larger grid sizes cause more simulated toughness, matching MPM trend where higher particle density raises material stiffness.
    }
	\label{fig:Appendix_grid_size}
\end{figure}

Specifically, we conducted hyperparameter analysis over the following parameter combinations: confidence $\in \{0.75,0.85,0.95\}$, max particle number $\in \{10,50,100\}$ and grid size $\in \{16,32,50\}$.
As shown in Fig.~\ref{fig:Appndix_confidence_and_max_particles}, we primarily investigated the influence of confidence and max particle number on the simulation results. The experiments show that increasing either parameter leads to more frequent bounces of the basketball after impact, indicating enhanced elasticity. We attribute this behavior to the increased number of surface particles introduced by higher parameter values. Since the predicted Young’s modulus of the surface material is higher than that of the internal material, the elevated surface particle ratio results in an overall increase in elastic behavior.
Additionally, in the Fig.~\ref{fig:Appendix_grid_size}, we performed an ablation study on grid size. The results show that increasing the grid size leads to higher toughness during simulation. This observation aligns with the behavior of the MPM framework, where increased particle density typically contributes to greater material stiffness.

\begin{table*}[t]
\centering
\begin{tabularx}{\textwidth}{lXXX} 
\hline
Metric \textbackslash Material Group & Snow body & Scarf & Hat \\
\hline
Top-1 Material/Ratio             & Snow/100\%   & Fabric/74\%  & Felt/100\%   \\
Top-1 $E$ ($\log_{10}(Pa)$)/Ratio & 6/48\%       & 7.699/24\%   & 7.699/68\%   \\
Top-1 $\rho$ ($kg/m^3$)/Ratio    & 100/94\%     & 1500/64\%    & 200/78\%     \\
Top-1 $\nu$/Ratio                & 0.3/82\%     & 0.4/68\%     & 0.3/74\%     \\
$E$ Mean/Median                  & 5.738/5.699  & 8.219/7.699  & 7.917/7.699  \\
$\rho$ Mean/Median               & 113/100      & 1369/1500    & 346/200      \\
$\nu$ Mean/Median                & 0.287/0.3    & 0.374/0.4    & 0.307/0.3    \\
\hline
\end{tabularx}
\caption{Statistics of 50 MLLMs estimations on the Snowman example. Reported metrics include Top-1 predicted material, 
Young’s modulus ($E$, in $\log_{10}(Pa)$), density ($\rho$, in $kg/m^3$), and Poisson’s ratio ($\nu$), along with mean and median values across multiple runs.}
\label{tab:Appendix_stats}
\end{table*}

\begin{table*}[t]
\centering
\begin{tabularx}{\textwidth}{lXXX} 
\hline
Examples \textbackslash MLLMs & GPT-4o & GPT-o3 & Gemini 2.5 Pro \\
\hline
Carnation (petal/stem/pot) & 6.00 / 7.176 / 9.301 & 6.00 / 7.301 / 9.176 & 6.00 / 7.00 / 9.176 \\
Basketball                 & 7.00             & 7.00             & 7.00          \\
\hline
\end{tabularx}
\caption{Comparison of Young’s modulus ($E$, in $\log_{10}(Pa)$) predicted by different MLLMs on the same objects.}
\label{tab:Appendix_mllms}
\end{table*}

\subsection{Reliability Analysis of MLLMs}
\label{B.5}
We conducted an experiment on the Snowman example (Fig.~\ref{fig:Editable}), performing 50 independent material property estimation, and reported the statistical results in Tab.~\ref{tab:Appendix_stats}. The results show that the material predictions from the MLLM do not follow a uniform random distribution, but instead exhibit strong preference patterns—indicating that the model tends to predict certain material properties more frequently and confidently.
We also compared the material indentification across different MLLMs for the same object, and summarized the predicted Young’s modulus values for each surface material in Tab.~\ref{tab:Appendix_mllms}. The results show a high degree of consistency across models. This inter-model agreement suggests that MLLMs may have already internalized shared knowledge about material properties during pretraining, reflecting real-world material semantics to some extent.
As demonstrated in all experiments in this paper, the 4D content generated based on MLLM-estimated material properties exhibits plausible visual dynamics and strong physical consistency. These results collectively indicate that MLLMs can serve as effective material property estimators, offering reasonably accurate predictions that support downstream physical simulation tasks.

\section{Preliminaries}
\label{C}
\subsection{3D Gaussian Splatting}
Unlike NeRF ~\cite{NeRF, Mip-NeRF, Instant-NeRF}, which uses implicit representations, 3D Gaussian Splatting (3DGS) employs explicit representations through a set of anisotropic Gaussian kernels~\cite{3DGS}, which can be interpreted as particles in space, enables physical simulations to simulate the particles' motion. The explicit representations also enable superior reconstruction quality and faster rendering speed. Specifically, 3DGS represents the 3D scene using a set of anisotropic Gaussian kernels defined by center position (mean) $\mathbf{x} \in \mathbb{R}^3$, covariance matrix $\mathbf{\Sigma} \in \mathbb{R}^7$, opacity $\alpha \in \mathbb{R}$, and color $\mathbf{c} \in \mathbb{R}^{3}$.
3D Gaussian kernel's spatial distribution can be expressed by:
\begin{equation}
    G(\mathbf{x}) = e^{-{\cfrac{1}{2}}(\mathbf{x})^T \mathbf{\Sigma}^{-1} (\mathbf{x})}.
\end{equation}
During the rendering process, the 3D Gaussians will be projected onto the image plane as 2D Gaussians. Sequentially, the pixel color $\mathbf{C}(\mathbf{\mathbf{r}})$ rendered from ray $\mathbf{r}$ can be computed by point-based volume rendering technique:
\begin{equation}
    \mathbf{C}(\mathbf{r}) = \sum_{i \in \mathcal{N}}\mathbf{c}_i\sigma_i\prod_{j=1}^i(1-\sigma_j),
\end{equation}
where $\sigma_i = \alpha_iG(\mathbf{x}_i)$, and $\mathcal{N}$ denotes the number of Gaussian kernel along ray $\mathbf{r}$. 
Because of this explicit representation, tracking the state of the Gaussian kernels becomes straightforward.
\subsection{Material Point Method}
\label{C.2}
Continuum mechanics studies the deformation and motion behavior of materials under forces. 
Motion is typically represented by the deformation map $\mathbf{x} = \phi(\mathbf{X}, t)$,
which maps from the undeformed material space $\omega^0$ to the deformed world space $\omega^t$.
The deformation gradient $\mathbf{F} = \frac{\partial \phi}{\partial \mathbf{X}} (\mathbf{X}, t)$ describes how the material deforms locally. 
MPM is a simulation method that combines Lagrangian particles with Eulerian grids and has demonstrated its ability to simulate various materials. 
In MPM, each particle $p$ carries various physical properties $\boldsymbol{\theta}_p^t$ at time step $t$, including mass $m_p$, density $\rho_p$, volume $V_p$, Young's modulus $E_p$, Poisson's ratio $\nu_p$, velocity $\mathbf{v}_p^t$, deformation gradient $\mathbf{F}_p^t$ and velocity gradient $\mathbf{C}_p^t$. The grid $i$ is used for computing intermediate results.
MPM operates within a loop that includes particle-to-grid (P2G) transfer, grid operations, and grid-to-particle (G2P) transfer.
In the particle-to-grid (P2G) stage, MPM transfers momentum and mass from particles to grids:
\begin{align}
(m\mathbf{v})_i^{t+1} &= \sum_p w_{ip} \left[ m_p \mathbf{v}_p^t + m_p \mathbf{C}_p^t (\mathbf{x}_i - \mathbf{x}_p^t) \right], \\
m_i^{t+1} &= \sum_p w_{ip}m_p,
\end{align}
where $w_{ip}$ is the B-spline kernel that measures the distance between particle $p$ and grid $i$. After P2G stage, we perform grid operations:
\begin{align}
\mathbf{v}_i^t &= (m \mathbf{v}_i)^t/{m_i^t} \\
\mathbf{f}_{i,in}^t &= - \sum_p \mathbf{P} (\mathbf{F}_p)(\mathbf{F}_p)^T \nabla w_{ip} \mathbf{v}_p^0 \\
\mathbf{v}_i^{t+1} &= \mathbf{v}_i^t + \Delta t \left( \mathbf{f}_{i,in} + \mathbf{f}_{ex} \right)/m_i \\
\mathbf{v}_i^{t+1} &= BC(\mathbf{v}_i^{t+1})
\end{align}

where BC refers to boundary condition. Then we transfer the results back to particles in the grid-to-particle (G2P) stage:
\begin{align}
\mathbf{v}_p^{t+1} &= \sum_i w_{ip} \mathbf{v}_i^{n+1}, \\
\mathbf{x}_p^{t+1} &= \mathbf{x}_p^t + \Delta t \mathbf{v}_p^{t+1}.
\end{align}
The velocity $\mathbf{v}_p^{t+1}$ and position $\mathbf{x}_p^{t+1}$ are updated using semi-implicit Euler method. Then, we update velocity gradient $\mathbf{C}_p^{t+1}$ and deformation gradient $\mathbf{F}_p^{t+1}$:
\begin{align}
\mathbf{C}_p^{t+1} &= \frac{4}{\Delta \mathbf{x}^2} \sum_i w_{ip} \mathbf{v}_i^{t+1} (\mathbf{x}_i - \mathbf{x}_p^t)^T, \\
\mathbf{F}_p^{t+1} &= \left( \mathbf{I} + \Delta t \mathbf{C}_p^{t+1} \right) \mathbf{F}_p^t. 
\end{align}
By following these three stages, we complete a simulation step.
\subsection{External Forces.}
In physical simulations, external forces—such as gravity—directly influence the motion and deformation of a continuum system.
In this paper, we apply two types of external forces. The first type directly modifies the particle’s velocity, such as setting the velocity of all particles to simulate the translation of the continuum. The second type indirectly affects the particle’s velocity by applying forces $\mathbf{f}$. For example, gravity can simulate the continuum’s fall. Given a force $\mathbf{f}$, we apply Newton’s second law and time integration to compute the particle’s velocity at the next time step:
\begin{equation}
    \mathbf{a}_p^{t} = \frac{\mathbf{f}}{m_p^t},
    \quad
    \mathbf{v}_p^{t+1} = \mathbf{v}_p^{t} + \mathbf{a}_p^t \Delta t,
\end{equation}
where $\mathbf{a}_p^t$ denotes the acceleration of particle $p$ at time step $t$, and $\Delta t$ denotes the time interval between time step $t$ and $t+1$.
By adjusting the external forces, we can control the motion and deformation of the object, thereby controlling the dynamics of the 4D content.
Additionally, since material grouping for 3D objects has been implemented in Sec. \ref{sec:Material Grouping}, we can precisely apply external forces to any material part of the object without relying on manual region selection, achieving user-friendly interaction and fine-grained control.


\end{document}